\documentclass[11pt]{article}

\usepackage[final]{acl}

\usepackage{times}
\usepackage{latexsym}

\usepackage[T1]{fontenc}

\usepackage[utf8]{inputenc}

\usepackage{microtype}

\usepackage{inconsolata}

\usepackage{graphicx}

\usepackage{graphicx}
\usepackage{placeins}
\usepackage{amsfonts}
\usepackage[ruled,vlined,linesnumbered]{algorithm2e}
\SetKw{KwDownTo}{down to}
\usepackage{xcolor}
\DontPrintSemicolon

\SetAlgoSkip{smallskip}

\title{Term-Centric Hierarchy Induction from Heterogeneous Corpora}

\author{Elena Senger$^{1,2}$ \quad Yuri Campbell$^{2}$ \quad  Jan-Peter Bergmann $^{2}$ \quad Rob van der Goot$^{3}$ \quad Barbara Plank$^{1}$ \\[1.5ex]
$^1$MaiNLP, Center for Information and Language Processing, LMU Munich, Germany \\
$^2$Fraunhofer Institute for Systems and Innovation Research ISI, Germany \\
$^3$Department of Computer Science, IT University of Copenhagen, Denmark \\
\texttt{elena.senger@cis.lmu.de, robv@itu.dk, b.plank@lmu.de} \\
\texttt{\{yuri.campbell,jan-peter.bergmann\}@isi.fraunhofer.de,}
}

\begin{document}
\maketitle
\begin{abstract}
Organizing knowledge from diverse text sources into interpretable hierarchies is crucial for tasks such as policy analysis, innovation monitoring, and exploratory domain mapping. Existing taxonomy induction methods typically rely on document-level representations that capture entire documents rather than the specific domain concepts relevant for knowledge organization, limiting their ability to generalize across heterogeneous sources.
We propose a term-centric framework for inducing hierarchical taxonomies from heterogeneous corpora that scales to massive document collections. Our approach maps documents from diverse sources into a shared representation space using automatic term extraction, enabling robust cross-source alignment. Based on these representations, we construct interpretable hierarchies that integrate domain priors with data-driven clustering.
Experiments on a novel English and German multi-source benchmark of over one million documents demonstrate that our method improves cross-source coherence and hierarchy quality over text- and summary-based baselines. A case study on German regional innovation analysis further demonstrates its practical utility for technology landscape mapping. 
\end{abstract}

\begin{figure*}
  \centering
  \includegraphics[width=1.0\linewidth]{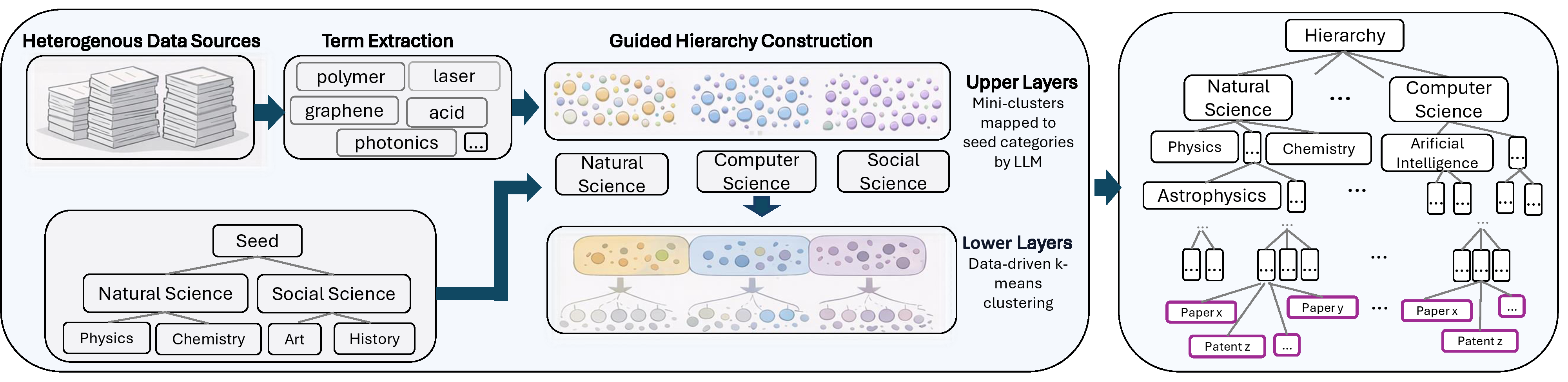}
  \caption{Overview of the TERMNET framework. Documents from heterogeneous sources are first mapped to a shared representation via automatic term extraction. The resulting embeddings are organized using a seed-guided hierarchical clustering procedure: predefined seed categories (representing broad scientific and technological domains) initialize the top levels of the hierarchy, which are then expanded in a data-driven manner.}
  \label{fig:TERMNET}
\end{figure*}

\section{Introduction}

Hierarchical representations organize large text corpora into interpretable, multi-level structures. Such hierarchies facilitate exploratory search, domain mapping, and trend analysis by enabling users to navigate from broad thematic areas to fine-grained topics. Recent clustering- and LLM-based approaches have shown promise for taxonomy induction from scientific literature \cite{zhu-etal-2025-context, katz-etal-2024-knowledge, oarga2024scientific, gao2025}. 
In practice, analysing complex domains often requires integrating heterogeneous sources reflecting different contexts. Tasks such as policy foresight, regional innovation analysis, and domain-specific knowledge discovery rely on synthesizing evidence from multiple data sources, for example to identify emerging technologies or monitor strategic priorities \citep{polchar2024foresight, hakiman2022innovation}.
Heterogeneous data sources pose two main challenges for hierarchy induction: 1) they differ in style and structure. For example, scientific papers emphasize methods and findings, patents focus on technical claims, and funding records describe strategic objectives. 
Standard document embeddings may therefore reflect source boundaries rather than thematic structure.
2) data-driven clustering follows the empirical corpus rather than the true structure of the domain. Since methods such as K-Means allocate more centroids to high-density or high-variance regions \citep{manning2008irbook}, sampling and coverage biases can lead to finer partitions where documents are abundant while under-represented areas may be merged or fragmented \citep{ester1996dbscan, mcinnes2017hdbscan}.

To address these challenges, we propose \textbf{TERMNET}, a scalable term-centric framework for inducing hierarchical taxonomies from heterogeneous corpora (Figure~\ref{fig:TERMNET}). Unlike prior taxonomy induction methods that rely on raw--document or summary embeddings, TERMNET maps documents into a shared semantic space using automatic term extraction, reducing source-specific style effects. 
Based on these representations, we construct hierarchies through a clustering process that integrates domain priors with data-driven signals, producing human-interpretable and domain-balanced taxonomy structures.
We evaluate TERMNET on our newly introduced large-scale multi-source benchmark containing over one million English and German documents. It outperforms raw-text and summary baselines in clustering quality, cross-source integration, and human interpretability, as validated by automatic and human evaluation. A policy-oriented case study further demonstrates the practical utility of the induced hierarchies.
Our main contributions are:
\begin{itemize}   \setlength\itemsep{0cm}

    \item  We introduce \textbf{TERMNET}, a scalable term-centric framework for hierarchy induction from heterogeneous corpora.
    
    \item We propose an evaluation protocol for multi-source hierarchy induction, including source entropy and intruder detection, and perform large-scale automatic and human evaluations.
    
    \item We release a multi-source benchmark of over one million documents spanning publications, patents, and funding records to support research on heterogeneous knowledge organization.
\end{itemize}

\section{Related Work}

Research on hierarchy induction has traditionally focused on pattern-based hypernym extraction~\citep[e.g.][]{hearst-1992-automatic, shwartz-etal-2016-improving, panchenko-etal-2016-taxi} and distributional or clustering-based methods that organize semantically related concepts or documents into hierarchical structures~\citep[e.g.][]{wang_2013, Liu_taxo_2012, mimno_2007}. More recently, LLMs have been applied either as standalone approaches or integrated into traditional pipelines~\citep[e.g.][]{gao2025, zhu-etal-2025-context, katz-etal-2024-knowledge}.

\subsection{LLM-Enhanced Hierarchy Induction}
Research on hierarchy induction has largely focused on single-source corpora. One early multi-source example, \citet{Zhu.2013},constructs topic hierarchies from blogs, community question-answering sites, and Twitter, but targets narrow topics and small-scale user-generated content. We extend this direction to large-scale institutional data sources and domain-level hierarchies.

\begin{table*}
\centering
\small
\resizebox{2\columnwidth}{!}{
\begin{tabular}{lcccl}
\hline
\textbf{Origin} & \textbf{raw text} & \textbf{Summary} & \textbf{Terms} & \textbf{Example} \\

\hline
\textit{openalex} & 197.12 & 71.62 & 50.34 & Marine Anthropogenic Litter; This book describes how ...\\
\textit{uspto}    & 107.80 & 62.57 & 21.38 & a canister vacuum cleaner comprising a main body ... \\
\textit{horizon}  & 274.56 & 80.54 & 67.01 & Redwave: the future of blood pressure measurement ... \\
\textit{foekat}   & 16.30  & 59.77 & 6.20 & Steigerung der Energieeffizienz beim Hochofenbetrieb ... \\
\hline
\textbf{Average} & 177.82 & 67.54 & 36.51 & --- \\
\hline
\end{tabular}
}
\caption{Mean text, summary, and parsed response lengths and example sentences per origin.}
\label{tab:mean_lengths}
\end{table*}

The work most closely related to ours is SCYCHIC \citep{gao2025}, which organizes scientific abstracts into multi-level hierarchies by combining embedding-based K-Means clustering with selective LLM-based summarization. A key insight is that decomposing papers into contribution types yields more coherent structures than treating each paper as a single-topic entity.
\citet{oarga2024scientific} leverage LLMs for zero-shot ontology and knowledge graph generation from scientific literature by prompting models to extract vocabulary, infer hierarchical category structures, and extract relations in an end-to-end manner, showing effectiveness in domain-specific settings, e.g. chemistry.
\citet{zhu-etal-2025-context} encode papers along multiple semantic aspects (e.g., methodology, data, evaluation) and cluster each summarized aspect with a probabilistic embedding-based model, followed by a dynamic search to ensure consistent cluster assignments when building a taxonomy.
\citet{katz-etal-2024-knowledge} introduce an LLM-guided framework that organizes scientific query results (a few thousand papers) into two-level hierarchies. Their system first embeds and clusters retrieved papers using Gaussian Mixture Models, followed by LLM-based naming, filtering, and grouping for exploratory browsing.

These methods operate on single-source scientific corpora and rely on document-level representations, such as raw text, summaries, or aspect-based reformulations.Summaries may preserve source-specific style or hallucinate content.
Therefore, we propose to use automatic terms extraction for hierarchy induction.

\subsection{Seed-Guided Hierarchy Induction}

Seed-guided hierarchy construction expands a small initial hierarchy using corpus evidence \citep{shen-etal-2025-unified}.
Early approaches rely on embedding-based methods that recursively organize concepts or attach new ones to existing nodes ~\citep[e.g.][]{taxogen_2018,taxocom_2022,Huang_2020}. 
More recent work leverages LLMs for seed-guided hierarchy construction. For instance, TAXOINSTRUCT \citep{shen-etal-2025-unified} uses instruction-tuned LLMs to generate sibling entities and infer parent relations, and other approaches iteratively extend seed hierarchies via prompting strategies \citep{gao2025}.
We use use seed hierarchies in a cross-source setting, combining seed-guided category assignment for the upper layers with data-driven clustering for the lower layers, to approximate the conceptual structure of the domain rather than the empirical distribution of individual datasets.

\section{Source Data}

To evaluate our approach, we construct a multi-source benchmark combining scientific publications, patents, and public research funding records. These sources capture complementary stages of the innovation pipeline: scientific knowledge production, technological protection, and publicly funded research activity. The dataset focuses primarily on German-affiliated organizations and contains documents in both German and English.
The corpus integrates four major data sources. It contains 578,335 publication abstracts from OpenAlex (\textit{openalex}) \citep{priem2022openalexfullyopenindexscholarly}, 353,043 patent abstracts from the USPTO corpus (\textit{uspto}) \citep{Li.2018}, 12,979 project descriptions from EU framework programs Horizon~2020 \citep{cordis_h2020_2015} and Horizon Europe (\textit{horizon}) \citep{cordis_horizon_europe_2022}, and 100,655 project descriptions from FöKAT (\textit{foekat}), which catalogs research projects funded by the German Federal Government \citep{fokat_2026}. The resulting dataset comprises 1,044,977 documents.

The sources differ substantially in their linguistic characteristics and document structure. Publication and patent abstracts typically contain well-structured descriptions of research contributions and technological inventions, whereas funding records are shorter and often contain administrative or program-specific terminology. Combining these heterogeneous sources provides a challenging benchmark for methods aiming to capture technological and scientific topics across institutional contexts.
Table~\ref{tab:mean_lengths} shows representative example documents and summary statistics for each source. Details on data retrieval, filtering criteria and licensing are provided in Appendix~\ref{app:preprocessing}.

\section{Methods}

\subsection{Problem Formulation}

Let $D = \{d_1, \dots, d_N\}$ denote a heterogeneous corpus consisting of documents from multiple sources. Each document may differ in structure, style, and length depending on its source.
Our goal is to induce a hierarchical taxonomy $\mathcal{H} = (\mathcal{V}, \mathcal{E})$ over the documents in $D$, where $\mathcal{V}$ represents the set of named taxonomy nodes, also referred to as categories, and $\mathcal{E}$ represents the unique parent--child relations between them.
Each node has a unique parent, except for the root
$v_0\in\mathcal{V}$, which has none.
The hierarchy organizes documents into increasingly specific categories. Each document is associated with a unique path from the root to a leaf node. For a node $v$, we use $D_v$ to denote the documents that pass through $v$ along this path, and
$\mathcal{C}_v$ to denote
its direct children.
The resulting hierarchy should satisfy two objectives: (i)
semantically coherent clusters, in which sibling nodes are thematically distinct and documents within a node share a common topic; and (ii) balanced domain coverage, such that the hierarchy reflects the breadth of the technological and scientific landscape rather than the frequency distribution of the corpus. We further assume access to domain priors $\mathcal{H}_p$ that define coarse-grained categories in the upper layers of $\mathcal{H}$. These priors may guide the hierarchy construction but do not fully determine its structure.

\subsection{TERMNET}
We propose TERMNET a term-centric framework for inducing scalable and interpretable hierarchies from heterogeneous corpora (Figure~\ref{fig:TERMNET}, Algorithm~\ref{algo:pseudo}). The key motivations are i) to abstract away from source-specific linguistic conventions and document structures by representing documents through their technological key-terms ii) to ensure interpretability with a guided hierarchy construction procedure that balances domain priors with data-driven discovery.

\paragraph{Term-centric Representation}
Documents from different sources may follow distinct stylistic conventions, which can cause document representations to reflect source identity rather than thematic structure. We address this by representing documents through their domain-specific terms—words or phrases denoting defined concepts within a specialized domain.
We use DiSTER, a fine-tuned model for cross-domain term extraction \citep{senger-etal-2025-crossing}, to identify concepts, methods, materials, and technologies in each document. The extracted terms are concatenated and embedded to form the document representation.
This maps heterogeneous texts into a shared representation space based on domain-specific terms. This encourages documents referring to similar concepts to align while reducing source-specific linguistic variation.

\paragraph{Hierarchy Guidance}
We construct the hierarchy using a hybrid strategy that combines domain priors with recursive clustering. The upper levels are initialized using predefined seed categories representing broad scientific domains, forming $\mathcal{H}_p$.
First, all documents are assigned to the root node $v_0$ (Alg.~\ref{alg:TERMNET}, line 2). 
Then, for each node $v$ with direct children,
we apply K-Means over term-based embeddings of the documents $D_v$ to obtain fine-grained clusters, where the number of clusters $k$ is determined as a multiple $\alpha$ of the number of seed categories under node $v$, that is $k = \alpha |\mathcal{C}_v|$.
Each cluster is then assigned to one of the existing children of $v$ or to a newly created child using zero-shot LLM classification (Alg.~\ref{alg:TERMNET}, lines 5–14).
For this decision, the LLM receives representative keywords and documents retrieved using class-TF–IDF \citep{Grootendorst2022BERTopicNT}. 
New nodes are created only if the cluster exceeds a minimum relative size $s_{\min}$ and if the cosine distance (dissimilarity) between the proposed label's embedding and the closest sibling label's embedding exceeds a threshold $\tau$.
$s_{\min}$ and $\tau$ aim to balance the guidance's depth granularity and suppress sibling explosion due to LLMs eagerness in creating new categories.

After reaching the seed depth $P$, the hierarchy is expanded in a fully data-driven manner using recursive top-down K-Means clustering (Alg.~\ref{alg:TERMNET}, lines 15–18). For each node, the number of children is chosen proportional to its document count while remaining within predefined bounds $(B_m, B_M)$. This allows dense nodes to be partitioned more finely while maintaining comparable granularity across hierarchy levels. Overall, the hybrid strategy preserves interpretability at higher levels while enabling fine-grained organization at lower levels.

\begin{algorithm}[t]
\caption{TERMNET algorithm}
\label{algo:pseudo}
\footnotesize
\SetAlgoNlRelativeSize{-1}
\SetInd{0.4em}{0.8em}
\label{alg:TERMNET}

\KwIn{Documents $\{d_1,\ldots,d_n\}$,
hierarchy guidance $\mathcal{H}_p$ with depth $P$, total layers $L$, cluster multiplier $\alpha$, size threshold $s_{\min}$, similarity threshold $\tau$, branching bounds $B_m,B_M$}

\textbf{Initialization:} Extract domain-specific terms for each $d_i$ and compute embedding $r_i$\;

Add all documents $d_i$ to the root node $v_0$\;
\For{$l = 0$ \KwTo $P - 1 $}{
    \For{each node $v$ in $\mathcal{V}_{l}$}
    {
        $k \leftarrow \alpha \cdot |\mathcal{C}_v|$ \;
        Create $k$ fine-grained clusters $\mathcal{M}$ using K-Means \;
        \For{Fine-grained cluster $m \in \mathcal{M}$}{
            Retrieve representative keywords and documents\;
            Zero-shot classify cluster $m$ into children of $v$\;
            \eIf{LLM proposes new category \textbf{and} $|m| \ge s_{\min}|D_v|$ \textbf{and} dissimilarity $> \tau$}{
              Create new child node $w\in\mathcal{C}_v$\;
                Assign documents of $m$ to $w$\;
            }{
              Assign documents of $m$ to selected child of $v$\;
            }
        }
    }
}
\For{$l = P$ \KwTo $L-1$}{
    \For{each node $v$ in $\mathcal{V}_{l}$}{
        Cluster document representations of $v$ into $k \in \{B_m, B_m+1 \cdots, B_M\}$ clusters\;
        Add every cluster as child of $v$\;
    }
}

\Return{Hierarchical structure}

\end{algorithm}

\subsection{Baselines}
We compare against hierarchical k-means clustering and a recent embedding taxonomy induction approach. Algorithms are provided in Appendix~\ref{app:algorithms}.

\paragraph{Recursive K-Means}
We implement recursive K-Means as a hierarchical clustering approach. Starting from a global partition, clusters are recursively subdivided in a breadth-first manner until the target granularity is reached. To ensure scalability, we apply Mini-Batch K-Means \citep{sulley_K-Means_2010} for clusters exceeding 10,000 documents and standard K-Means \citep{zis-MacQueen1967Some} otherwise. This approach produces a strictly nested hierarchy in which each document is assigned to a path defined by successive centroid refinements. We apply this baseline to three document representations: raw text, document summaries, and extracted domain-specific terms.
\paragraph{SCYCHIC} 
Due to compute feasibility, we adapt the SCYCHIC framework \citep{gao2025} to our large-scale corpus by using smaller embedding and summarization models (see Section~\ref{exp_setup}). To address token constraints, in the bottom-up phase, we summarize only a representative subset of documents per cluster selected using Maximal Marginal Relevance (MMR), balancing similarity to the cluster centroid with semantic diversity.

\section{Experimental Design}
\subsection{Setup}
\label{exp_setup}
All methods use Qwen3-Embedding-0.6B \citep{zhang2025qwen3embeddingadvancingtext} to generate document embeddings. For document summarization, we employ Meta-Llama-3-8B-Instruct~\citep{grattafiori2024llama3herdmodels}, and for autmatic term extraction we use DiSTER with instruction-tuned Meta-Llama-3-8B-Instruct~\citep{grattafiori2024llama3herdmodels} following~\citet{senger-etal-2025-crossing}.

The hierarchy guidance $\mathcal{H}_{p}$ is fixed with two category layers curated by domain experts and inspired by the ASJC classification \citep{ASJC-Scopus}. Fine-grained clustering uses a multiplier $\alpha = 50$ that determines the number of clusters per node. The creation of new categories during guidance is controlled by a minimum relative cluster size $s_{\min} = 1\%$ and a cosine similarity threshold $\tau = 0.12$ to prevent redundant labels. For the data-driven hierarchy expansion phase, the branching factor is bounded by $B_m=3$ and $B_M=6$. The final hierarchy depth is fixed to $L=4$, resulting in hierarchies with approximately $k=\{6, 40, 180, 680\}$ clusters from top to bottom. This configuration was selected in consultation with domain experts to support the downstream case study and analyses (Section~\ref{sec:case}). However, the method itself is not restricted to a fixed hierarchy depth or number of clusters.

\begin{table*}
\small
\centering
\resizebox{2\columnwidth}{!}{
\begin{tabular}{lccccccc}
\hline
Method & Precision & Diversity & Entropy  & Intruder & Structure & Validity & Composition \\
\hline

K-Means (raw)
& 53.28 & \textbf{99.79} & 24.50  & 84.30  & 2.0 & 2.5 & 2.5 \\

K-Means (summary)
& 47.56 & 99.41 & 23.98 & \underline{91.25}  & 3.5 & 2.0 & 4.0 \\

K-Means (terms)
& 61.12 & 99.75 &  30.00  & 86.42  & 3.0 & 1.5 & 4.0 \\

SCYCHIC
& 60.52 & 99.73 & 23.48  & 79.02 & \underline{4.0} & \underline{3.0} & 4.0 \\ 

\hline

TERMNET wo. terms
& 66.39 & 99.57 &  \underline{33.11}  & 69.94  & \underline{4.0} & 2.5 & \underline{5.0} \\

\hline

TERMNET
& \textbf{67.70} & 99.22 & \textbf{36.70} & \textbf{94.44} & \textbf{4.5} & \textbf{4.5} & \textbf{5.0} \\

\hline
\end{tabular}
}
\caption{Automatic metrics and human evaluation scores across the different hierarchy induction approaches. TERMNET wo. terms differs only in the input representation from the full TERMNET approach by using raw text instead of extracted terms.}
\label{tab:topic_metrics_dataset2_human}
\end{table*}

\subsection{Evaluation}
We evaluate the hierarchies with unsupervised, supervised, and human-centered metrics to assess clustering quality, cross-source integration, structure and usability.

\paragraph{Unsupervised Metrics}
We measure lexical cluster distinctiveness at each hierarchy level using topic \textit{diversity}. Topic diversity is measured using Inverted Rank-Biased Overlap~\citep{webber2010similarity, terragni2021word} over the top $10$ representative terms per cluster. 
Cross-source integration is measured with \textit{source entropy} for each cluster based on the distribution of document origins. Higher entropy indicates a more balanced source mixture, while lower entropy reflects source dominance. 
Perfect balance is unattainable due to skewedness in source-sizes and also not necessarily desirable, hence we comparatively evaluate reductions in source-driven fragmentation using entropy.
Preliminary experiments showed that clustering raw documents produced source-separated clusters with low entropy; therefore, higher entropy indicates improved integration. The composition of clusters from different sources is further validated through human evaluation.

\paragraph{Supervised Metrics}
To evaluate high-level thematic alignment, we manually construct a gold standard from source-specific classifications (e.g., OpenAlex fields and IPC codes for patents) and map these categories to induced top-level clusters. Documents are considered correctly assigned if their original categories correspond to the mapped cluster. For example, publications labeled with \textit{General Immunology and Microbiology} and \textit{Health Informatics} are expected to appear in the corresponding \textit{Life Sciences} or \textit{Health Sciences} clusters. Since we only have these labels for a subset of the data, we can only measure precision.

To evaluate the semantic coherence, we design an intruder detection task inspired by \citet{bhatia-etal-2018-topic}. Given a parent cluster and its child clusters, along with one semantically similar intruder cluster drawn from a different parent at the same hierarchy depth, the task is to identify the non-matching cluster.
In a validation study we randomly sampled 60 parent clusters, balanced across hiarchy levels. For each instance, two human annotators independently identified the intruder cluster. Human inter-annotator agreement reached Cohen’s $\kappa = 0.73$, while 
intruder predictions from an LLM (Llama-3.3-70B-Instruct~\citep{grattafiori2024llama3herdmodels}) matched annotators with mean $\kappa = 0.76$ and higher mean precision, so we use it for large-scale evaluation and report macro-accuracy per layer.

\paragraph{Human Evaluation}
Following \citet{hu2025taxonomytreegenerationcitation} and \citet{zhu-etal-2025-context}, we evaluate hierarchies along three dimensions adapted to cross-data source hierarchy induction: \textit{Structure}, \textit{Validity} and  \textit{Composition}. \textit{Structure} assesses whether the hierarchy exhibits a clear progression from broad research areas to specific sub-technologies with consistent granularity across levels. \textit{Validity} evaluates alignment with experts’ understanding of how concepts are typically grouped and related. \textit{Composition} examines whether clusters reflect a plausible mix of data sources given the domain (e.g., publication- vs.\ patent- or project-intensive areas). Each dimension is rated on a five-point Likert scale (see Appendix~\ref{app:human_eval}). The mean inter-annotator agreement is Cohen’s $\kappa = 0.43$. \footnote{The moderate chance-corrected agreement scores ($\kappa = 0.43$, $\alpha = 0.57$) reflect minor calibration differences and low variation. While exact agreement is 44\%, 87\% of ratings differed by at most one Likert point. Both annotators consistently ranked TERMNET highest, indicating disagreement mainly affect absolute ratings rather than system ordering. For per-annotator see Appendix~\ref{app:likert}.}

\section{Results \& Analysis}

\subsection{Quantitative Analysis}

Table~\ref{tab:topic_metrics_dataset2_human} summarizes automatic metrics and human evaluation scores. Overall, \textsc{TERMNET} achieves the strongest performance across most metrics, obtaining the highest precision (67.70), intruder accuracy (94.44), entropy (36.70), and human evaluation scores. Topic diversity remains very high for all methods ($\approx99$), indicating that clusters are lexically distinct and largely non-overlapping regardless of the clustering strategy.

\begin{figure*}[t]
  \centering
  \includegraphics[width=0.95\linewidth]{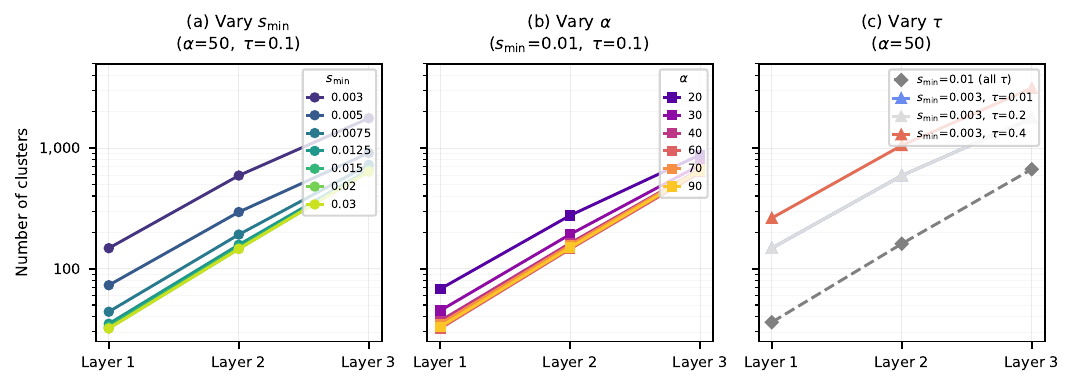}
  \caption{Sensitivity of the hierarchy shape to the three construction hyperparameters.}
  \label{fig:sensitivity}
\end{figure*}

\paragraph{Clustering Quality}
Among the baselines, K-Means with term representations achieves the highest precision (61.12), outperforming raw text (53.28) and summaries (47.56), while SCYCHIC reaches 60.52. This suggests that term-centric representations reduce source-specific linguistic variation and better capture technological concepts. 
Seed guidance further improves alignment, \textsc{TERMNET} without terms reaches 66.39 precision and the highest baseline entropy (33.11). The full model performs best overall, with 67.70 precision and 94.44 intruder accuracy. The large improvement in intruder accuracy compared to baselines (e.g., 84.30 for K-Means raw and 79.02 for SCYCHIC) indicates substantially higher semantic coherence of the generated hierarchy.

\paragraph{Structure and Validity}
Table~\ref{tab:topic_metrics_dataset2_human} shows that \textsc{TERMNET} obtains the highest expert ratings for Structure and Validity (both 4.5). The strongest baselines reach Structure scores of 4.0 and Validity scores of at most 3.0. Even without term extraction, \textsc{TERMNET} archives Structure = 4.0, indicating that hierarchy guidance alone already improves the organization of domains.

\paragraph{Cross-Source Integration}
Term representations substantially increase source entropy, indicating better integration of publications, patents, and funding records. K-Means with terms increases entropy to 30.00 compared to 24.50 for raw text. \textsc{TERMNET} further improves integration, achieving the highest entropy (36.70). The human \textit{Composition} of 5.0 confirms that it's clusters contain plausible source mixtures, rather than reflecting source-specific artifacts.

\subsection{Sensitivity Analysis and Practical Guidance}
\label{sec:sensitivity}
We analyse the three hyperparameters controlling hierarchy construction: the over-clustering multiplier $\alpha$, the minimum relative size $s_{\min}$ for creating guided nodes, and the similarity threshold $\tau$ for avoiding redundant labels. Figure~\ref{fig:sensitivity} shows how the parameters affect the hierarchy shape.
Reducing $s_{\min}$ from 0.03 to 0.003 increases cluster counts by about a factor of four, from 32 to 148 at layer~1 and from 639 to 1,768 at layer~3, as smaller candidate clusters survive the guided merging step. The effect saturates for $s_{\min} \geq 0.02$, where further increases no longer affect the hierarchy.
The seed multiplier $\alpha$ has a weaker but visible effect, smaller values create more clusters, while results stabilize for $\alpha \geq 60$ for the given $s_{\min}$ and  $\tau$.
The hyperparameter are cros-dependent.  At $s_{\min}=0.01$, sweeping $\tau$ yields identical cluster counts. In contrast at $s_{\min}=0.003$, $\tau=0.4$ nearly doubles the number of clusters compared to $\tau \leq 0.2$. Thus, $\tau$ only becomes active when the size threshold is permissive enough for many small candidate clusters to be considered for new labels. Practically, we therefore recommend choosing $s_{\min}$ first, using $\alpha$ as a secondary density control, and tuning $\tau$ only for very low $s_{\min}$.

\begin{table}[t]
\small
\centering

\begin{tabular}{lcc}
\hline
$(\alpha, s_{\min}, \tau)$ & Entropy & Intruder \% \\
\hline
$(20, 0.01, 0.10)$   & 46.16 & 96.01 \\
$(40, 0.003, 0.10)$  & 44.50 & 96.28 \\
$(50, 0.01, 0.16)$   & 47.20 & 94.09 \\
$(50, 0.01, 0.40)$   & 48.20 & 92.57 \\
$(70, 0.02, 0.10)$   & 48.64 & 95.11 \\
\hline
\end{tabular}
\caption{Evaluation scores for five sensitivity configurations.}
\label{tab:sensitivity_scores}
\end{table}

\begin{table*}
\centering
\resizebox{2\columnwidth}{!}{
\begin{tabular}{lcccccc}
\hline
\textbf{Cluster} & Example Keywords & \textit{openalex} & \textit{uspto} & \textit{horizon} & \textit{foekat} & Method  \\
\hline

Social Sciences And Humanities & [stakeholders, populist, mental] &  148.668 & 4.309  & 1.806 & 3.993 & K-Means (raw) \\
Materials Science And Advanced Technologies & [nanostructures,  superconducting] &  130.126 &  53.368  & 1.935 & 3.248 & K-Means (raw) \\
Engineering And Applied Technologies & [reconfigurable, exoskeletons] & 16.636 & 275.257  & 78 & 205 & K-Means (raw) \\
Biomedical And Molecular Sciences & [fibrosis, protein, myelin] & 140.317 & 11.017  & 1.810 & 3.526 & K-Means (raw) \\
Life Sciences & [protein,biomass, genome] & 103.866 & 5.595  & 1.558 & 2.586 & K-Means (raw) \\
Information And Communication Technologies & [5g, innovativ, autonome] & 38.722 & 3.462  & 5.792 & 87.097 & K-Means (raw) \\
\hline
Social Sciences And Humanities & [kant, acculturation, unemployment] &  55.764 & 192  & 671 & 6.140 & TERMNET \\
Natural Sciences & [polymerization, phonon, chiral] &  168.263 &  31.326  & 2.165 & 6.874 & TERMNET \\
Engineering & [epitaxial, actuator, doped] & 53.877 & 168.741  &  1.724 &  22.927 & TERMNET \\
Health Sciences & [microglia, ecmo, hfpef] & 139.286 & 16.937  & 1.696 & 7.793 & TERMNET \\
Life Sciences & [isoforms, vesicles, rnas] & 117.821 & 41.472  & 3.153 & 24.360 & TERMNET \\
IT and Computing & [virtualized, subcarriers, ldpc] & 43.324 & 94.340  &  3.570 &  32.561 & TERMNET \\
\hline
\end{tabular}
}
\caption{Top-level clusters and document counts per data source for the raw text K-Means baseline and \textsc{TERMNET}. Cluster names for the baseline were generated using an LLM. Keywords represent truncated examples of characteristic terms for each cluster. The distribution illustrates how TERMNET produces more balanced cross-source clusters compared to raw text clustering.}
\label{tab:top_layer_distribution}
\end{table*}

Table~\ref{tab:sensitivity_scores} supports this interpretation: across five configurations, intruder accuracy remains between 92.57 and 96.28, and normalized source entropy between 44.50 and 48.64. The parameters therefore control navigational granularity while leaving measured quality comparatively stable.

\subsection{Qualitative Analysis}
\begin{table*}
\centering
\resizebox{2\columnwidth}{!}{
\begin{tabular}{lcccccccc}
\hline
Method & Precision & Diversity & Entropy  & Intruder \% & Structure & Validity & Composition & Usefullness \\
\hline

K-Means (raw)
& 33.70 & \underline{99.81} & 22.91  & \textbf{98.01} & -- & -- & -- & -- \\

K-Means (summary)
& 37.71 & 99.13 & 18.07 & \underline{95.63} & -- & -- & -- & -- \\

K-Means (terms)
& 39.96 & 99.78 & \underline{28.02}  & 86.34 & \underline{3.5}  & \underline{4} & 2 & \underline{3.5}\\

SCYCHIC
& \underline{51.71} & 99.77 & 20.71 & 90.24 & 1  & 2 & \textbf{3} & 2\\

\hline

TERMNET
& \textbf{64.77} & \textbf{99.88} & \textbf{35.63} & 92.54 & \textbf{4.5}  & \textbf{4.5} & \textbf{3} & \textbf{4}\\

\hline
\end{tabular}
}
\caption{Automatic metrics and human evaluation scores for the proprietary dataset.}
\label{tab:topic_metrics_dataset1_human}
\end{table*}

\begin{table}
\centering
\small
\begin{tabular}{lccccc}
\hline
Method & Entropy & Accuracy \%\\
\hline
K-Means raw text  & 13.28  & 80 \\
K-Means summary  & 11.75 & 78 \\ 
K-Means terms   & 21.49 & \textbf{86} \\
\hline
\end{tabular}
\caption{Impact of document representations on source entropy and human-perceived accuracy.}
\label{tab:topic_metrics_layer0}

\end{table}
Analysis of intruder detection errors reveal subtle hierarchy limitations. \textsc{TERMNET} without terms performs worse at broad levels because some clusters include semantically distant subclusters. For example `Social and Community Development` was placed under `Life Sciences`, causing the model to select it instead of the true intruder, `Mathematics`.With term extraction, such social-science subclusters are absent from `Life Sciences`, and the true intruder 'Automotive Engineering' is identified correctly. Such errors illustrate that while the seed-guided hierarchy ensures broad structural coherence, precision at fine-grained levels benefits from term-centric representations.

We hypothesize that the high \textit{Composition} scores are largely driven by term extraction, \textit{Structure} and \textit{Validity} benefit from hierarchy guidance. The upper layers of the hierarchy are initialized using the seed hierarchy, resulting the upper layers which roughly resemble established scientific hierarchies. This familiar structure helps experts navigate the taxonomy. In contrast, purely data-driven approaches can deviate from established structures. For example, K-Means with summaries creates detailed engineering clusters at the top level but omits social sciences and humanities, and deeper clusters such as `Policy and Governance Studies` appear under `Life Sciences`, reflecting less coherent assignments without seed guidance.

Table~\ref{tab:top_layer_distribution} further illustrates differences in cluster composition. Raw text K-Means produces clusters that are strongly dominated by individual data sources. For example, 'Engineering and Applied Technologies' contains 275k USPTO documents and 205k FÖKAT records, while other sources are sparsely represented. In contrast, \textsc{TERMNET} produces more balanced clusters. For instance, the 'Engineering' and 'IT and Computing' clusters contain substantial contributions from both OpenAlex and USPTO documents, suggesting that documents from different sources referring to similar concepts are aligned more effectively.

These observations suggest that \textsc{TERMNET} balances hierarchy guidance with data-driven clustering, producing hierarchies that are interpretable at high levels while capturing fine-grained, semantically coherent subtopics.

\section{Case Study: Structural Change}
\label{sec:case}
To illustrate \textsc{TERMNET}'s utility for regional analysis, we provide induced hierarchies to geographers studying structurally weak German regions. This case study uses in-house data, substituting publicly available OpenAlex abstracts with Scopus abstracts for higher-quality metadata and PATSTAT replaces USPTO to focus on German affiliations. The \textit{horizon} and \textit{foekat} datasets are used as before. \textsc{TERMNET} achieves the highest scores on most  metrics(Table~\ref{tab:topic_metrics_dataset1_human}), demonstrating robust performance. For human evaluation, we added a \textit{Usefulness} dimension to capture the practical utility of the hierarchies for geographers. Expert annotations were limited to the most promising approaches due to resource constraints.

We further evaluate clustering quality and cross-source integration using K-Means with $k=650$ on three input types: raw text, summaries, and terms (Table~\ref{tab:topic_metrics_layer0}). Only a single clustering layer was generated. From each clustering, 50 documents were randomly sampled for blinded manual assessment. A geography expert judged whether each document was thematically consistent with its cluster, using the raw document text, representative keywords, representative cluster documents, and an LLM-generated cluster label. Term-centric representations achieved higher source entropy and human-perceived accuracy, indicating improved cross-source mixing and more coherent clusters.

\section{Conclusion}
We introduced \textsc{TERMNET}, a term-centric framework for inducing hierarchical taxonomies from heterogeneous corpora. By combining domain-specific term representations, seed-guided construction, and data-driven clustering, \textsc{TERMNET} produces interpretable hierarchies that align with established domain structures while integrating multiple data sources. Experiments show consistent gains in thematic alignment, semantic coherence, and cross-source integration across automatic metrics and expert evaluations.

\FloatBarrier

\section*{Limitations}
While TERMNET demonstrates strong performance in cross-source hierarchy induction, several limitations remain that offer directions for future research. First, the interpretability and structure of the upper hierarchy levels depend on the quality of the expert-provided seed categories. While this guidance improves alignment with established domain structures, poorly specified seeds may bias the resulting taxonomy. Second, our evaluation focuses on English and German corpora within scientific and technological domains. The generalizability of the approach to other languages, domains, or document genres remains an open question. Third, the current implementation constructs a static hierarchy from a fixed snapshot of documents. In real-world technology monitoring scenarios, corpora evolve continuously as new publications and patents appear. Supporting incremental updates and modelling the temporal evolution of hierarchical structures would therefore be an important extension for future research.

\section*{Ethical Considerations}

This work uses publicly available textual data sources and does not involve personal or sensitive information. Nevertheless, hierarchical representations learned from large corpora may reflect biases present in the underlying data, which could influence how topics or domains are structured and interpreted. Users should therefore treat automatically induced hierarchies as exploratory tools rather than authoritative representations of knowledge.

Large language models were used to assist with grammar correction, spelling, and refinement of the manuscript text. They were not used to generate experimental results, analyses, or scientific claims.

\bibliography{custom}

\appendix

\section{Details on the Dataset Creation and Licensing}
\label{app:preprocessing}

\subsection{Dataset Creation}
This appendix provides details on the data retrieval, filtering, and text construction steps applied to each source in the dataset.

\begin{itemize}
    \item OpenAlex \citep{priem2022openalexfullyopenindexscholarly}: We retrieved publications via the OpenAlex API with at least one German-affiliated author for the period 2015–2023. Only records with a DOI were retained. The text corpus was constructed by concatenating the \texttt{title} and \texttt{abstract} fields.
    
    \item USPTO \citep{Li.2018}: We used patent records from the USPTO dataset for the years 2014–2015. The textual content consists of the \texttt{Abstract} field.
    
    \item CORDIS: We used project data from Horizon~2020 (funded 2014-2020) \citep{cordis_h2020_2015} and Horizon Europe (funded 2021-2027) \citep{cordis_horizon_europe_2022}. The text corpus combines \texttt{ProjectTitle} and \texttt{ProjectObjective}.
    
    \item FöKAT \citep{fokat_2026}: We used funding records from 2014–2025, with the year determined by \texttt{Laufzeit von}. The text corpus consists of the \texttt{Thema} field. Compared to the other datasets, FOEKAT entries are substantially shorter and frequently contain funding-specific terminology or proper names of funding programs without descriptive context. To mitigate this, domain experts excluded selected categories from the FÖKAT taxonomy (\textit{Leistungsplansystematik}) and applied additional keyword-based filtering to remove highly specific funding instruments (e.g., \textit{Professorinnenprogramm}).
\end{itemize}

Across all sources, we additionally removed extremely short entries (less than five characters).
\subsection{Licensing}
The benchmark dataset aggregates information from United States Patent and Trademark Office (USPTO) patents (CC BY 4.0), OpenAlex publications (CC0), CORDIS project data (generally available under CC BY 4.0 / PSI reuse terms), and FöKAT records used with permission. To ensure license compatibility and proper attribution requirements across sources, the compiled benchmark dataset is released under the Creative Commons Attribution 4.0 (CC BY 4.0) license.

\section{Algorithms}
\label{app:algorithms}
In Algorithm \ref{alg:recursive_kmeans}, we present the classical approach to hierarchical clustering using recursive K-Means.
In Algorithm \ref{alg:scychic}, we reproduce the SCYCHIC algorithm, introduced by  \citet{gao2025}.

\begin{algorithm}[t]
\footnotesize
\SetAlgoNlRelativeSize{-1}
\SetInd{0.4em}{0.8em}
\caption{Recursive K-Means hierarchy construction}
\label{alg:recursive_kmeans}

\KwIn{Set of documents $\{d_i,\ldots,d_n\}$, number of layers $L$, target number of clusters $(k_1,\ldots,k_L)$}

\textbf{Initialization:} Embed each $d_i$

\For{$l = 1$ \KwTo $L$}{
    \eIf{$l = 1$}{
        Cluster all documents into $k_1$ clusters\;
    }{
        \For{each cluster $c$ from layer $l-1$}{
            Cluster documents of $c$ into $k_l$ subclusters\;
        }
    }
}

\Return{Hierarchical structure}

\end{algorithm}

\begin{algorithm}[t]
\caption{Scychic algorithm}
\footnotesize
\SetAlgoNlRelativeSize{-1}
\SetInd{0.4em}{0.8em}
\label{alg:scychic}

\KwIn{Set of documents $\{d_i,\ldots,d_n\}$, number of layers $L$, target number of clusters $(k_1,\ldots,k_L)$}

\textbf{Initialization:} Embed each $d_i$

\For{$l = 1$ \KwTo $\lfloor L/2 \rfloor$ \tcp*[r]{Top-down phase}}{
    \eIf{$l = 1$}{
        Cluster all documents into $k_1$ clusters\;
    }{
        \For{each cluster $c$ from layer $l-1$}{
            Cluster documents of $c$ into $k_l$ subclusters\;
        }
    }
}

\For{each cluster $\tau$ at level $\lfloor L/2 \rfloor$ \tcp*[r]{Bottom-up phase}}{
    \For{$l = L$ \KwDownTo $\lfloor L/2 \rfloor + 1$}{
        \eIf{$l = L$}{
            $E$ = \{ embeddings of documents within $\tau$ \} \;
        }{
            $E$ = \{ embeddings of summaries of cluster $l+1$ \} \;
        }
        Cluster $E$ into $k_l$ subclusters \;
        Generate summary for each subcluster
    }
}

\Return{Hierarchical structure}

\end{algorithm}

\section{Details on Human Evaluation}
\label{app:human_eval}
The four dimonsions for the Likert scale are:
\begin{itemize}
    \item \textbf{Structure:} Does the organization follow a clear logical hierarchy, transitioning from broad research areas to specific sub-technologies, or do the clusters maintain a consistent level of granularity on each layer?
    \item \textbf{Validity:}	Does the taxonomy align with experts’ understanding of the scientific and technological landscape, including how concepts are commonly grouped, related, or distinguished?
    \item \textbf{Composition:} Does the clusters reflect an appropriate and plausible composition of data sources given the nature of the technology (e.g., publication-dominated scientific fields versus patent- or project-intensive applied domains)?
    \item \textbf{Usefulness (Case Study):} How useful is the hierarchy for policy analysts or foresight practitioners seeking to monitor technological developments, compare signals across data sources, or identify emerging or underexplored areas?
\end{itemize}

The 1-5 Likert Scale is defined as: 
\begin{enumerate}
    \item \textbf{Completely inaccurate}, with significant factual errors or misrepresentations of the domain.
    \item \textbf{Mostly inaccurate}, capturing only a few correct facts but failing to represent the domain coherently.
    \item \textbf{Moderately accurate}, containing some factual correctness but missing important concepts or relationships.
    \item \textbf{Mostly accurate}, representing the domain well with minor factual inaccuracies or omissions.
    \item \textbf{Highly accurate}, thoroughly reflecting the domain's factual structure with no noticeable errors.
\end{enumerate}

\section{Per-Annotator Likert Scale Ratings}
\label{app:likert}
Tables~\ref{tab:ann1} and \ref{tab:ann2} present the per-annotator Likert-scale ratings. TERMNET consistently achieves the highest score, either independently or tied with other approaches, with one exception: for annotator~1 in the “Structure” dimension, where it scores one point below the top-rated approach.

\begin{table}[h!]
\centering
\resizebox{1\columnwidth}{!}{
\begin{tabular}{c|ccc}
\hline
Item & Structure & Validity & Composition \\
\hline
K-Means (raw) & 2 & 3 & 2 \\
K-Means (summary) & 4 & 2 & 5 \\
K-Means (terms) & 3 & 1 & 4 \\
SCYCHIC & 5 & 3 & 4 \\
TERMNET wo. terms & 5 & 2 & 5 \\
TERMNET  & 4 & 4 & 5 \\
\hline
\end{tabular}
}
\caption{Likert-scale ratings from Annotator 1}
\label{tab:ann1}
\end{table}

\begin{table}[h!]
\centering
\resizebox{1\columnwidth}{!}{
\begin{tabular}{c|ccc}
\hline
Item & Structure & Validity & Composition \\
\hline
K-Means (raw) & 2 & 2 & 3 \\
K-Means (summary) & 3 & 2 & 3 \\
K-Means (terms) & 3 & 2 & 4 \\
SCYCHIC & 3 & 3 & 4 \\
TERMNET wo. terms & 4 & 3 & 5 \\
TERMNET & 5 & 5 & 5 \\
\hline
\end{tabular}
}
\caption{Likert-scale ratings from Annotator 2}
\label{tab:ann2}
\end{table}

\end{document}